\documentclass[runningheads]{llncs}
\usepackage[T1]{fontenc}
\usepackage{graphicx}
\usepackage{hyperref}
\usepackage{amsmath} 
\usepackage{multirow}
\usepackage[numbers]{natbib}

\begin{document}
%

\title{CLRecogEye : Curriculum Learning towards exploiting convolution features for Dynamic Iris Recognition}
\titlerunning{CLRecogEye}
%
\author{Geetanjali Sharma\inst{1} \and
Gaurav Jaswal\inst{2} \and
Aditya Nigam\inst{1} \and
Raghavendra Ramachandra \inst{3}}
\authorrunning{Geetanjali Sharma et al.}
\institute{Indian Institute of Technology Mandi, India \and
Directorate of Forensic Services, Shimla, India \and
Norwegian University of Science and Technology (NTNU), Norway
\\}

\authorrunning{G. Sharma et al.}

%

%
\maketitle              
\begin{abstract}
Iris authentication algorithms have achieved impressive recognition performance, making them highly promising for real-world applications such as border control, citizen identification, and both criminal investigations and commercial systems. However, their robustness is still challenged by variations in rotation, scale, specular reflections, and defocus blur. In addition, most existing approaches rely on straightforward point-to-point comparisons, typically using cosine or L2 distance, without effectively leveraging the spatio-spatial-temporal structure of iris patterns. 
To address these limitations, we propose a novel and generalized matching pipeline that learns rich spatio-spatial-temporal representations of iris features. Our approach first splits each iris image along one dimension, generating a sequence of sub-images that serve as input to a 3D-CNN, enabling the network to capture both spatial and spatio-spatial-temporal cues. To further enhance the modeling of spatio-spatial-temporal feature dynamics, we train the model in curriculum manner. This design allows the network to embed temporal dependencies directly into the feature space, improving discriminability in the deep metric domain.
The framework is trained end-to-end with triplet and ArcFace loss in a curriculum manner, enforcing highly discriminative embeddings despite challenges like rotation, scale, reflections, and blur. This design yields a robust and generalizable solution for iris authentication.\textit{Github code}: \href{https://github.com/GeetanjaliGTZ/CLRecogEye
}{\texttt{https://github.com/GeetanjaliGTZ/CLRecogEye
}}.


\end{abstract}
\section{Introduction}
Contactless biometric traits such as iris \cite{wei2022towards, wei2022contextual, zhang2023secure}, face, fingerprint, palmprint, and forehead creases \cite{bharadwaj2022mobile, sharma2024fh} have garnered significant attention for personal identification due to the high distinctiveness of their spatial textures, shapes, and line patterns. Among these, iris recognition offers unique advantages: (a) the complex vascular texture of the iris exhibits variability even among identical twins, (b) its discriminative features remain visually stable over time, and (c) it demonstrates limited genetic heritability. These properties render iris recognition particularly robust under challenging conditions. Furthermore, unlike contact-based modalities, iris recognition ensures greater user convenience and mitigates hygiene concerns—an aspect that has gained increased relevance during the COVID-19 pandemic.
However, iris recognition has traditionally relied on constrained scenarios to acquire high-quality images. Near-infrared (NIR) iris recognition has been widely deployed in large-scale biometric systems, such as the Indian Aadhaar program, which enables both offline and online identity verification of residents using iris images. Recent research has explored solutions that enable iris acquisition in semi-controlled environments from a distance, even with only cooperative subjects. Additionally, efforts have been made to extend iris recognition to mobile devices using visible-wavelength illumination. The demonstrated utility of iris recognition has spurred increasing research into more accurate and robust iris matching algorithms.
\subsection{Open Challenges and Our Work}
Our objective diverges from conventional CNN-based biometric matching approaches (e.g.,~\cite{schroff2015facenet, deng2019arcface, thapar2019pvsnet}). Rather than developing a new recognition system from scratch, we aim to exploit the capability of 3D-CNNs to learn spatio-spatial-temporal representations from stacked iris sub-images. This allows us to address a fundamental question: \emph{Can CNNs automatically model the role of non-rigid motion in biometric matching?} Specifically, we investigate whether CNNs can capture motion-related cues in a manner consistent with human intuition under variations in pose, illumination, and acquisition quality. In contrast, prior CNN-based methods rely primarily on standard convolutional filters, which are limited in modeling such dynamics, while triplet-loss frameworks further complicate training due to the challenge of triplet sampling.  
To overcome these limitations, we propose a deep metric learning framework for robust iris matching across diverse datasets and imaging conditions. Our architecture integrates a 3D-CNN with triplet and ArcFace loss to extract highly discriminative spatio-spatial-temporal features invariant to rotation, scale, and illumination. This is achieved by dividing each iris image into overlapping patches, stacking them, and processing consecutive sequences through a 3D-CNN equipped with rectangular filters to enrich intra-patch dynamics.  
Furthermore, motivated by the large variability in iris image quality and acquisition conditions, we incorporate \emph{curriculum learning} to gradually expose the network to increasingly challenging samples. This strategy facilitates stable optimization, improves generalization, and enables the model to capture inter-patch dependencies and long-range temporal patterns more effectively. Finally, the framework is trained end-to-end with triplet \& ArcFace loss, yielding discriminative embeddings that generalize well across datasets.  \textbf{Contributions.} The key contributions of our work are:  
\begin{itemize}
    \item To the best of our knowledge, this is the first study to explicitly model non-rigid motion in iris recognition by treating an image as a stack of overlapping patches.
    \item We propose a lightweight CNN architecture with rectangular filters and 3D convolutions to learn rich intra-patch spatio-spatial-temporal features.
    \item We present a deep metric learning framework with extended Triplet + ArcFace loss, which improves feature discriminability and generalization across iris datasets.
\end{itemize}

\section{Literature survey}

Early iris recognition relied primarily on hand-crafted feature descriptors, most notably the Gabor phase-quadrant representation known as IrisCode \cite{daugman2009iris}. Subsequent efforts explored alternative descriptors, including block-wise spectral analysis with DCT coefficients \cite{monro2007dct}, multi-lobe differential filters \cite{sun2008ordinal}, texture analysis \cite{ma2003personal}, phase-based approaches \cite{miyazawa2008effective}, zero-crossings \cite{sanchez2002iris}, and intensity variation analysis \cite{ma2004local}.
With the advent of deep learning, convolutional neural networks (CNNs) emerged as the dominant paradigm in biometric recognition due to their strong feature learning capability. Early CNN-based iris models include DeepIris \cite{liu2016deepiris}, which employed a 9-layer architecture for nonlinear similarity mapping, and DeepIrisNet \cite{gangwar2016deepirisnet}, a baseline deep learning approach for multi-spectrum iris recognition. Pre-trained CNNs were also leveraged, with features extracted for classification using SVMs \cite{nguyen2017iris}, while capsule networks incorporating pre-trained submodules were investigated in \cite{zhao2019deep}. Despite these advances, CNN-based methods often struggle to generalize under noise, reflections, blur, and occlusions, which exacerbate intra-class variation.
To mitigate these challenges, deep metric learning has been widely adopted. Inspired by DeepFace \cite{taigman2014deepface} and FaceNet \cite{schroff2015facenet}, CNN-based iris models incorporated triplet loss \cite{zhao2017towards}, later enhanced with residual learning and dilated kernels \cite{wang2019toward}. Other studies trained triplet-loss networks on non-normalized images with hard-triplet mining \cite{ahmad2019thirdeye}, while ComplexIrisNet \cite{nguyen2017iris} employed Gabor complex images with DenseNet and extended triplet loss. Dynamic strategies, such as adaptive-margin hard-negative mining \cite{thapar2019pvsnet}, further improved accuracy, yet triplet-loss frameworks remained limited by small mini-batch constraints and convergence issues.
More recent work has introduced uncertainty-aware and context-based formulations. Iris Recognition by Learning Uncertain Factors \cite{wei2022towards} proposed uncertainty embedding (UE) and uncertainty-guided curriculum learning (UGCL) to explicitly model acquisition noise and improve generalization. By treating iris images as probabilistic distributions, UE captured both identity and uncertainty features, while UGCL facilitated more robust training. Complementary to this, Contextual Measures (CM) \cite{wei2022contextual} modeled correlations between iris regions using parallel branches for global and local context, achieving further improvements in accuracy and efficiency.
Finally, beyond iris, alternative biometric modalities such as Photoplethysmography (PPG) have been investigated. For instance, \cite{zhang2023secure} proposed a secure authentication framework for IoT healthcare, leveraging Homomorphic Random Forests (HRF) on encrypted PPG features, with dimensionality reduction via PCA \cite{abdi2010principal} and feature selection using mRMR \cite{peng2005feature}, achieving both high accuracy and low error rates.

\section{Matching 3D Iris Images in Wild}
In this section, we describe the preprocessing of iris images, encompassing segmentation, normalization, and masking to emphasize the distinctive textural patterns of the iris. Furthermore, we detail the input representation for the base model, wherein the original 2D images are transformed into a 4D tensor to facilitate the extraction of both local and global features.

\subsection{Iris image Preprocessing} 
Raw iris images from datasets cannot be used directly, as they contain multiple regions—including the eyelid, pupil, sclera, and iris—of which only the iris region is relevant for recognition. Effective preprocessing is therefore essential to isolate this informative region. In particular, the iris, which occupies the lower portion of the image, exhibits distinctive textural patterns that are critical to achieve reliable and accurate recognition performance.
\begin{figure}[h]
\centering
\includegraphics[width=0.8\textwidth]{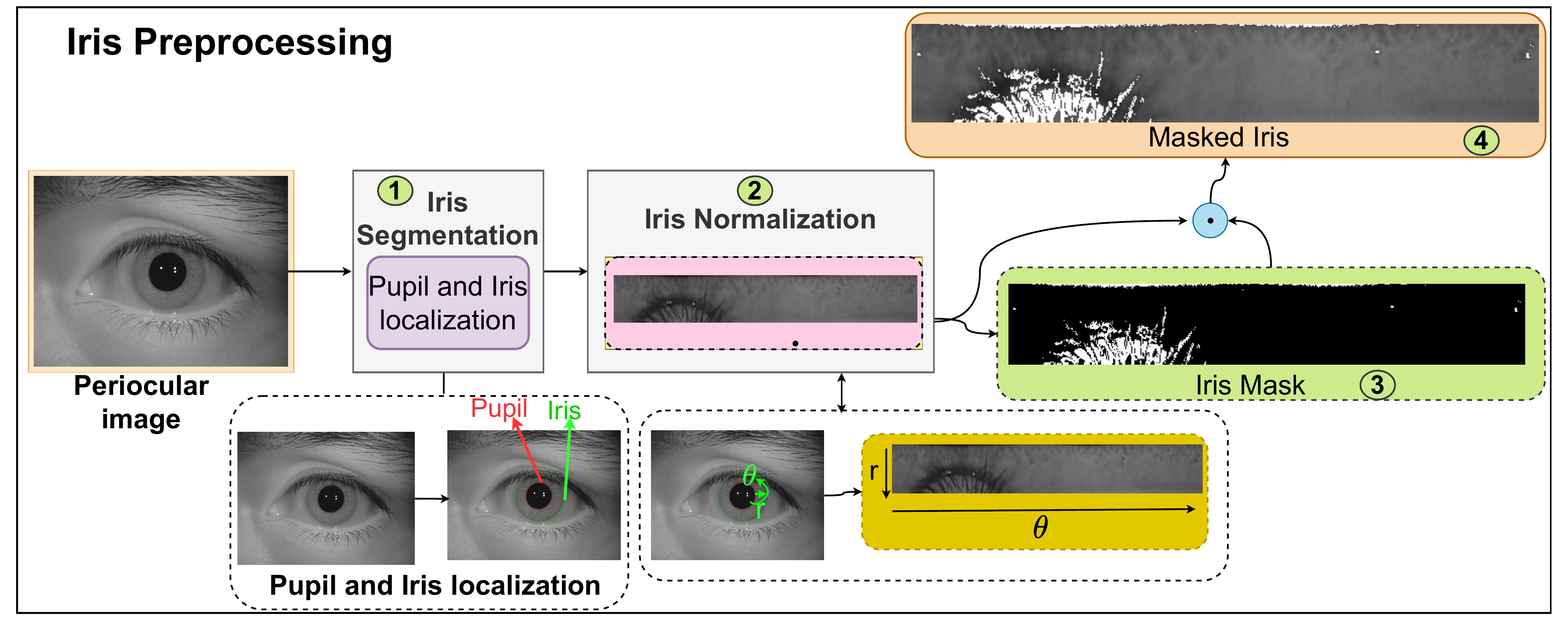}
\caption{Steps of input Iris image preprocessing: 1) Segment the pupil and iris from the periocular iris input images 2) Normalize the segmented iris are using daugman rubbersheet model 3) Masked iris : Pixel wise multiplication of iris mask and iris image}
\label{fig:IP1}
\end{figure}
The preprocessing stage comprises three key steps: (i) iris localization, (ii) iris segmentation, and (iii) iris normalization \cite{daugman2009iris}. Localization and segmentation are performed using PixSegNet \cite{jha2020pixisegnet}, which detects the iris and pupil regions by estimating their centers and radius. Segmentation produces a binary mask, where the iris is represented by white pixels and the background by black pixels. This mask is subsequently used to generate normalized iris images that retain essential textural information while discarding irrelevant background, as illustrated in Figure~\ref{fig:IP1}.
Raw images in iris datasets exhibit varying resolutions (e.g., $620 \times 480$, $320 \times 480$). After preprocessing, each iris is represented in a compact, strip-like format that is significantly smaller than the original image, as shown in Figure~\ref{fig:IP1}. These normalized iris images are then used for subsequent recognition tasks.

\subsection{Image composition}
\begin{figure}[h]
\centering
\includegraphics[width=0.8\textwidth]{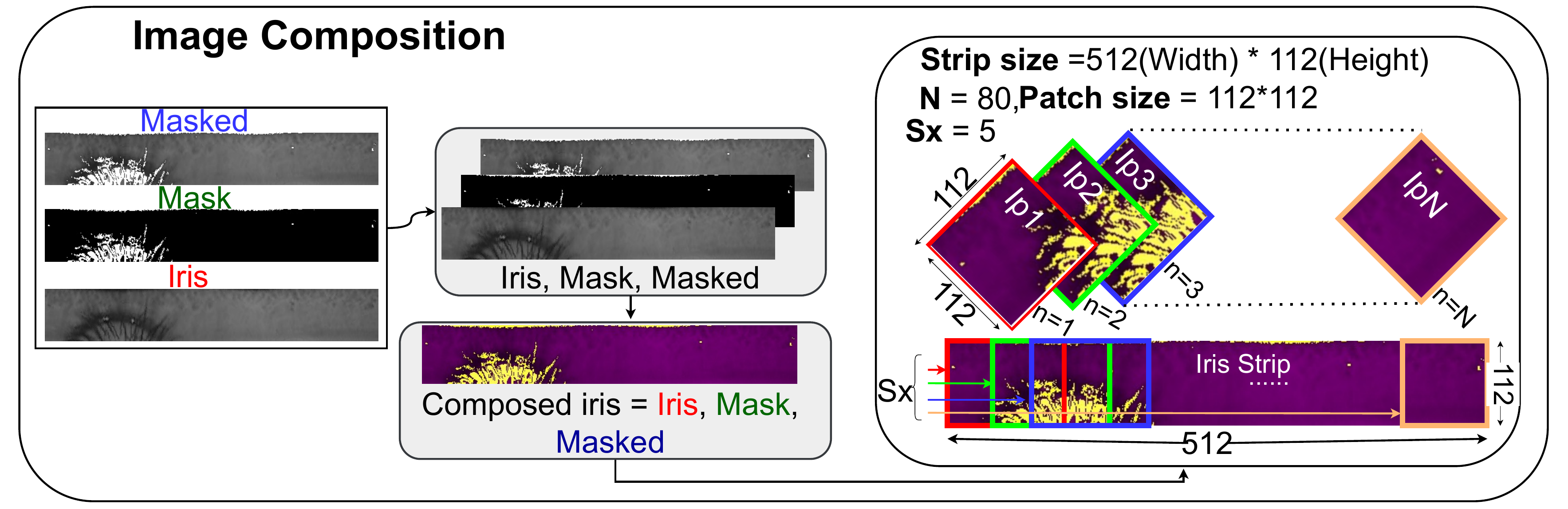}
\caption{Image composition and patch generation. A grayscale iris image is converted into a three-channel representation (iris, mask, masked iris), normalized to $112 \times 512$, and partitioned into $N = 80$ overlapping patches ($112 \times 112$) with a vertical stride of $S_y = 5$ and $S_x = 0$.}
\label{fig:IP2}
\end{figure}

For each experiment, we start with a two-dimensional grayscale iris image, which is transformed into a three-channel representation. Specifically, the first channel encodes the original iris image, the second contains the corresponding segmentation mask, and the third represents the masked iris image. Together, these channels form a comprehensive three-channel input representation.
The normalized iris image, with a size of $112 \times 512$, is subsequently partitioned into 80 overlapping patches of size $112 \times 112$. A vertical stride of 5 pixels ($S_y = 5$) is applied, while no stride is applied along the horizontal direction ($S_x = 0$), as illustrated in Figure~\ref{fig:IP2}. Therefore, each patch has dimensions $112 \times 112 \times 3$, and stacking all 80 patches yields a four-dimensional tensor of size $80 \times 112 \times 112 \times 3$.
This transformation effectively expands the original 2D input into a 4D representation, allowing the network to exploit both global and local iris information by jointly modeling spatial and spatio-spatial-temporal dependencies. The stacked patches, as shown in Figure~\ref{fig:IP2}, clearly illustrate the key steps of this process.

\subsection{Proposed Curriculum Learning Framework : CLRecogEye}
In this section, our proposed framework is structured into two primary components: the \textit{Backbone}, which employs the Modified Inflated 3D (I3D) model, and the \textit{Head}, which integrates a full connected layer trained with arcface loss and triplet loss curriculum.

\textbf{Modified Inflated 3D:} 
To effectively capture both fine-grained iris textures and their variations within and across patches, we adopt the Two-Stream I3D network \cite{carreira2017quo}. Originally introduced for video understanding. I3D extends 2D convolutional kernels into 3D, enabling hierarchical spatio-spatial-temporal feature learning within the Inception architecture (Figure \ref{fig:IP2}). This design allows simultaneous modeling of subtle iris patterns and their local (within patch) and global(across patches) dependencies. While I3D introduces additional parameters compared to 2D CNNs, the temporal modeling (across patches) capability substantially enhances discriminative power. For adaptation to iris recognition, we reconfigure the initial blocks to match input iris resolution and replace square kernels with rectangular ones for more efficient iris feature extraction. To satisfy the network’s channel requirements and enrich representation, each sample is encoded as a three-channel input composed of the raw iris image, its segmentation mask, and the masked iris (Figure~\ref{fig:IP2}). Training is supervised using categorical cross-entropy loss, promoting robust discriminative feature learning and alignment with class labels.
To support this, the backbone network adopts an Inception-based 3D CNN architecture (Figure~\ref{fig:IP2}), which captures rich spatio-spatial-temporal features from iris dataset. The network comprises 3D convolutional, batch normalization, and max-pooling layers. The initial convolutional layer operates on four-dimensional tensors (channels, frames, width, height), using $(N \times N \times N)$ filters with stride $2$ and same padding to capture both spatial and temporal dependencies, followed by 3D max-pooling for down-sampling. To enrich the feature space, Inception blocks employ parallel branches with heterogeneous filter sizes—$(1 \times 3 \times 7)$, $(7 \times 7 \times 3)$, $(1 \times 1 \times 1)$, and $(3 \times 3 \times 3)$—enabling multiscale representation learning. This design progressively captures discriminative iris patterns from local details to higher-level abstractions. At the classification stage, 3D global average pooling condenses each feature map to a scalar, forming a compact embedding vector. This vector is optimized with ArcFace loss, followed by softmax activation, to produce class probabilities across $783$ subjects. By combining spatio-temporal modeling, multi-scale feature extraction, and margin-based metric learning, the model produces highly discriminative embeddings for robust iris recognition.

\subsection{Iris Curriculum Learning Framework}

To enhance discriminative feature learning and ensure stable convergence, we propose a \textit{curriculum-based optimization strategy} for our iris recognition network, denoted as $CLRecogEye$. Instead of relying solely on a single loss function, we employ an \textit{alternating loss training scheme} that cycles between \textbf{Triplet Loss} and \textbf{ArcFace Loss} during different stages of optimization. Specifically, training begins with Triplet Loss to achieve initial convergence and establish well-separated inter-class boundaries. After a few epochs, once the network learns coarse discriminative embeddings, the optimization switches to ArcFace Loss to refine intra-class compactness by enforcing angular margin constraints in the embedding space. This alternation is repeated for three full cycles, yielding improved convergence stability and superior performance compared to either loss function used independently.

\textbf{(i) Triplet Loss:} 
For iris recognition, the triplet loss is employed to learn highly discriminative embeddings through adaptive hard and soft sample mining. Each triplet consists of an anchor ($x_a$), a positive sample ($x_p$) from the same subject, and a negative sample ($x_n$) from a different subject. The objective is to ensure that the distance between the anchor–positive pair is smaller than that of the anchor–negative pair by at least a margin $m$, formulated as:
\begin{equation}
\mathcal{L}_{\text{triplet}} = \frac{1}{N} \sum_{i=1}^{N} 
\max\left(0, 
\|f(x_a^i) - f(x_p^i)\|_2^2 - 
\|f(x_a^i) - f(x_n^i)\|_2^2 + m 
\right)
\end{equation}
where $f(\cdot)$ denotes the embedding function and $N$ is the batch size. In our framework, the margin $m$ starts at 0.5 and gradually increases up to 1.5; if semi-hard triplets in a batch fall below 10\% of anchor-positive pairs for three consecutive batches, the margin is incremented by 0.05 to maintain training difficulty.

\textbf{(ii) ArcFace Loss: } 
Cross-entropy loss, commonly used for feature classification, effectively separates classes in closed-set tasks but often produces scattered embeddings with poor intra-class compactness and limited generalization. 
To address this, the loss of ArcFace\cite{deng2019arcface} introduces an additive angular margin $(m)$ to improve intra-class compactness and inter-class separability.
\begin{equation}
L = -\frac{1}{N} \sum_{i=1}^{N} 
\log \frac{
\exp \left(s \cdot \cos(\theta_{y_i} + m)\right)
}{
\exp \left(s \cdot \cos(\theta_{y_i} + m)\right)
+ \sum_{j=1, j \neq y_i}^{n} \exp \left(s \cdot \cos \theta_j \right)
}
\end{equation}

where $\theta_{y_i}$ is the angle between the normalized feature $\mathbf{x}i$ and the corresponding class weight $\mathbf{W}{y_i}$, $m$ is the angular margin (typically $m=0.4$), and $s$ is the scaling factor. ArcFace normalizes features and weights and adds an angular margin to enforce compact, discriminative embeddings for open-set recognition.

\textbf{Theoretical Justification:} 
Triplet Loss encourages \textit{relative distance-based discrimination} by maximizing the margin between anchor-positive and anchor-negative pairs, which is particularly effective during the early training stages when class boundaries are underdeveloped. However, Triplet Loss does not explicitly promote intra-class compactness or efficient utilization of the embedding space. Conversely, ArcFace Loss introduces an \textit{additive angular margin} on the normalized hypersphere, ensuring embeddings from the same class form tightly clustered manifolds while maintaining inter-class separability through angular constraints.
By alternating between these two complementary objectives, the network benefits from both \textit{coarse-grained inter-class separation} (via Triplet Loss) and \textit{fine-grained intra-class compaction} (via ArcFace Loss). This dynamic optimization strategy helps navigate the non-convex loss landscape more effectively, prevents premature convergence to suboptimal minima, and promotes a well-structured embedding manifold that captures both global separability and local compactness. Empirically, this curriculum results in smoother convergence, faster stabilization, and enhanced generalization across cross-domain iris recognition scenarios.

\subsection{Final Matching Function: Cosine Similarity}
We have finally used Cosine Similarity (\(S_{C}\)) metric to measure the similarity between the learned feature vectors. It evaluates the angle between two vectors to determine whether they point in the same direction. The purpose of cosine similarity is to quantify the relationship between gallery and probe features, as expressed in Equation~\ref{eq:cosine}. The cosine similarity between vectors $\mathbf{Probe}$ and $\mathbf{Gallery}$ is defined as:

\begin{equation}
S_{C}(\text{Probe}, \text{Gallery}) := \cos (\theta) = \frac{\mathbf{Probe} \cdot \mathbf{Gallery}}{\|\mathbf{Probe}\|\|\mathbf{Gallery}\|}
\label{eq:cosine}
\end{equation}

\section{Experimental Evaluation}
Extensive experiments were conducted to evaluate the proposed approach under different settings. Results are reported on NIR \& VIS Iris datasets collected under varying imaging conditions using standard evaluation metrics. 


\subsection{Data-sets and Evaluation Metrics}
In this work, we conduct experiments on seven representative and publicly available iris databases: CASIA-Iris Lamp V3, CASIA-Iris Lamp V4, Blue Iris (BI), and Dark Iris (DI), where the latter includes captures from iPhone (P1), Nokia (P2), and SGS5 (P3). All experiments were performed according to standard protocols. The benchmark details are as follows:

\begin{itemize}
\item \textbf{Dataset Split and Protocol:} 
We follow a subject-independent protocol, where images for each subject are evenly divided into two disjoint subsets: \textit{gallery} and \textit{probe}. This 50-50 split ensures non-overlapping sets for fair evaluation.

\item \textbf{Metrics:} Performance is evaluated using the correct recognition rate (CRR), equal error rate (EER) and the decidability index (DI). Additionally, Receiver Operating Characteristic (ROC) curves are plotted based on False Rejection Rate (FRR) and False Acceptance Rate (FAR) for detailed comparisons.
\item \textbf{Evaluation Protocol:} The proposed model is trained on 50\% of the CASIA-V3 training dataset and evaluated on all other benchmark datasets.  

\end{itemize}
    

\subsection{Experiment setting}
Our experimental framework was implemented in \texttt{Python} using the TensorFlow backend and executed on a Linux workstation equipped with an NVIDIA GPU. To enhance the discriminative capacity of the learned representations, we employed the triplet loss and then ArcFace loss with an additive angular margin of $m=0.2$ and a scaling factor of $s=30$. The model was trained and evaluated on the \textbf{CASIA-Iris Lamp-V3} dataset, which comprises 783 subjects. Each subject includes 20 iris samples—10 used for training (\textit{gallery}) and 10 for evaluation (\textit{probe})—resulting in 78,300 genuine and 612,306 impostor pairs for performance analysis.
The proposed framework employs the Inflated 3D ConvNet (I3D) backbone, adapted for iris feature extraction using rectangular filters. A curriculum training strategy first applies triplet loss for feature compactness, then ArcFace loss for class separation, ensuring stable optimization and robust iris embeddings.

\section{Results and Discussion:}  
Table~\ref{tab:results} presents the quantitative results of our approach. On the CASIA-Lamp-V3 dataset, the proposed framework achieves a closed-set EER of \textbf{0.35\%}, with TMR values above \textbf{99\%} at both 0.1\% and 0.01\% FMR, confirming the strong discriminative capability of the model. In the open-set setting, two distinct configurations were evaluated: the first achieved an EER of \textbf{0.70\%}, a CRR of \textbf{99.73\%}, and a DI of \textbf{2.05}; the second further reduced the EER to \textbf{0.23\%}, with a CRR of \textbf{99.87\%} and a DI of \textbf{3.07}, demonstrating superior generalization. 
On the CASIA-Lamp-V4 dataset, the model attained EER values of \textbf{1.53\%} (left eye) and \textbf{1.69\%} (right eye), with corresponding CRRs of \textbf{99.40\%} and \textbf{99.18\%}, and DI values of \textbf{1.46} and \textbf{1.27}. These results highlight robust performance across both subsets, though left-eye samples consistently yield slightly better separability.  
Performance on cross-sensor datasets (BI and DI) is naturally lower due to challenging imaging conditions (e.g., DI-P3 with an EER of \textbf{21.37\%}), yet the framework still maintains consistent recognition trends. Overall, the combination of Modified I3D, triplet loss, ArcFace loss, and cosine similarity ensures state-of-the-art performance in controlled environments and provides strong generalization under open-set and cross-device conditions.
\begin{table*}[htbp]
    \centering
    \caption{Iris recognition results for CASIA Lamp-V3 \& V4, Blue and Dark iris Dataset}
    \begin{tabular}{lcccccc}
    \hline
    \multirow{2}{*}{\textbf{Dataset}} & \multirow{2}{*}{\textbf{EER (\%)}} & \multicolumn{2}{c}{\textbf{TMR (\%) @ FMR (\%) =}} & \multirow{2}{*}{\textbf{Genuine}} & \multirow{2}{*}{\textbf{Imposter}}\\ \cline{3-4}
     &  & \textbf{0.1 (\%)} & \textbf{0.01 (\%)} \\ \hline
    \multicolumn{5}{c}{\textbf{CASIA-Iris Lamp-v3 Dataset}} \\ \hline
    Closed-set & \textbf{0.35} & \textbf{99.62} & \textbf{99.59}& 78300& 61230600 \\ \hline
    Open-set (ES) & 0.70 & 98.67 & 97.43& 148580& 60996000 \\ \hline
    Open-set (OS) &  0.23 & 99.73 & 99.64& 148580& 60996000 \\ \hline
    \multicolumn{5}{c}{\textbf{CASIA-Iris Lamp-V4 Dataset}} \\ \hline
    Open-set(LE) &  \textbf{1.42} & \textbf{97.73} & \textbf{97.05}& \textbf{14744}& \textbf{6296705} \\ \hline
    Open-set(RE) &  1.69 & 97.08 & 95.08& 152364& 64980606 \\ \hline
    \multicolumn{5}{c}{\textbf{BlueIris (BI) and Dark Iris(DI) Dataset}} \\ \hline
    BI-P1 & 13.74 & 45.71 & 64.49& 490 & 36566 \\ \hline
    BI-P2 &  20.05&  25.15& 12.57& 344& 28396 \\ \hline
    DI-P1 &  12.57 &  32.75 &  19.78& 14450& 2139106 \\ \hline
    DI-P2 &  22.61&  20.98& 10.12& 45728& 6171814 \\ \hline
    DI-P3 &  21.37&  27.67& 17.76& 5204& 423166 \\ \hline
    \end{tabular}    
    \label{tab:results}
\end{table*}

\begin{figure}[!ht]
\centering
\includegraphics[width=0.8\textwidth]{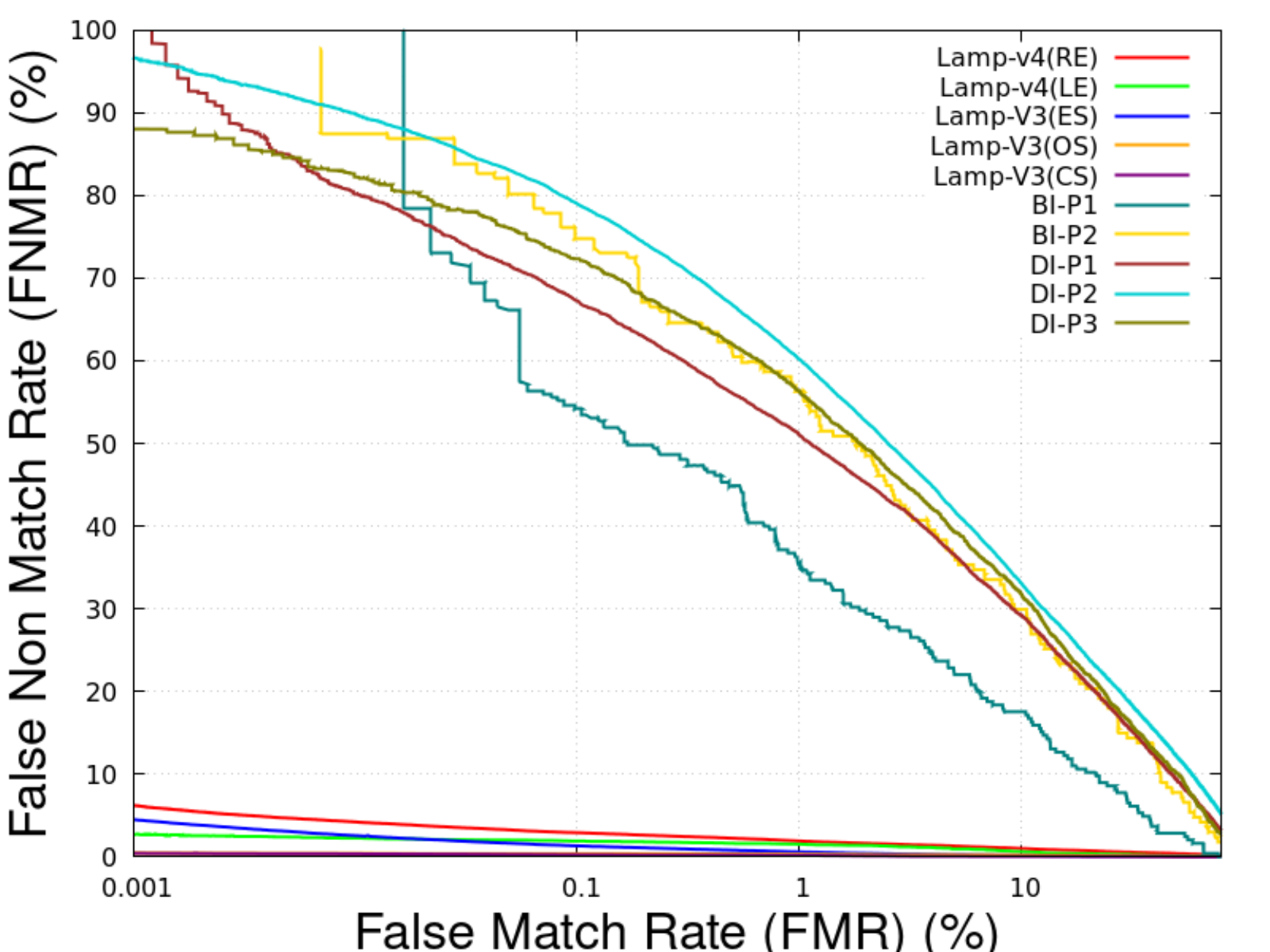}
\caption{DET curve of the proposed system on the Lamp-V3, Lamp-V4, BI-P1, BI-P2, DI-P1, DI-P2 and DI-P3 showing the relationship b/w False Rejection Rate and False Acceptance Rate.}
\label{fig:det_curve}
\end{figure}

In addition to the tabulated metrics, Figure~\ref{fig:det_curve} shows the Detection Error Trade-off (DET) curves for the CASIA-Lamp, Blue-Iris (BI), and Dark-Iris (DI) datasets. The DET curve visualizes the relationship between False Match Rate (FMR) and False Non-Match Rate (FNMR), enabling direct performance comparison across decision thresholds. The proposed framework achieves consistently low FNMR at very low FMR on CASIA-Lamp-V3 and V4, demonstrating strong discriminative embeddings. In contrast, BI and DI datasets, particularly DI-P2 and DI-P3, exhibit higher error rates due to cross-sensor and cross-environment variations. Nonetheless, the overall downward trend of the curves indicates robust separation between genuine and impostor pairs. These results align with Table~\ref{tab:results}, confirming that combining ArcFace loss, triplet loss, and cosine similarity with the I3D backbone produces highly discriminative and generalizable iris representations, especially under controlled imaging conditions.

\section{Conclusion}
In this work, we presented a curriculum-guided iris recognition framework that integrates the I3D architecture with triplet and ArcFace loss functions to jointly learn spatio-spatial-temporal features and discriminative representations. Experiments conducted on the CASIA-Lamp, Blue-Iris, and Dark-Iris datasets demonstrate that our approach achieves competitive recognition performance, effectively capturing both local texture variations and global structural dependencies. The incorporation of curriculum learning further stabilizes optimization and enhances generalization across varying illumination and sensor conditions.
Despite strong performance, the framework remains sensitive to imperfect segmentation and imbalanced samples. Future work will focus on adaptive curriculum strategies and multimodal fusion. We also plan to develop lightweight models for robust real-time deployment.
\bibliographystyle{IEEEtran}
\bibliography{ref}{}
%
%
%




\end{document}